\documentclass[10pt,twocolumn,letterpaper]{article}
\PassOptionsToPackage{table}{xcolor}

\usepackage[pagenumbers]{cvpr}
\usepackage{booktabs}
\usepackage{booktabs,tabularx,array,multirow}

\definecolor{DTURuntimeGreen}{HTML}{4C9673}
\usepackage{booktabs,tabularx,array}

\newcommand{\DTUcell}[3]{%
  \cellcolor{DTURuntimeGreen!#1!white}%
  \ensuremath{#2\!\rightarrow\!#3}%
}

\definecolor{cvprblue}{rgb}{0.21,0.49,0.74}
\usepackage[pagebackref,breaklinks,colorlinks,allcolors=cvprblue]{hyperref}

\title{Dynamic Task Update Benchmark (DTU-Bench): Evaluating Runtime Task Adaptation in Robotic Manipulation }

\author{Yanran Jiang$^{1}$, C\'ecile Paris$^{1}$, Dana Kuli\'c$^{1,2}$, and Leimin Tian$^{1}$\\
$^{1}$CSIRO Technology, Australia\\
$^{2}$Monash University, Australia}

\begin{document}
\maketitle

\begin{abstract}
Vision-language-action (VLA) policies perform well on fixed-instruction robotic manipulation benchmarks, but their ability to adapt to instruction changes during execution remains understudied. We introduce the Dynamic Task Update Benchmark (DTU-Bench), evaluating 40 task revisions across seven update types and three execution stages using LIBERO, a simulated robotic manipulation benchmark. We compare performance when a new instruction is given before execution starts versus during execution, and examine whether the new task instructions are followed. Across 13 representative VLA configurations, average task success is 95.8\% on the original object-manipulation tasks, compared with only 1.6\% when the instruction changes which object to manipulate during execution. Controlled diagnostics show that policies often complete the original task before the updated one, and that their responses depend on the robot's physical state. We further test whether a simple scripted controller that releases the grasped object and moves the arm back helps robots respond to instruction changes. This improves success from 2\% to 38\% for $\pi_{0.5}$ and from 8\% to 48\% for DM0.5, but can cause unwanted object motion. These findings reveal limitations that fixed-instruction benchmarks and task-success metrics alone can overlook in evaluating robotic manipulation. Project page: \url{https://dtu-bench.github.io/}.
\end{abstract}

\begin{figure*}
    \centering
    \includegraphics[width=1\linewidth]{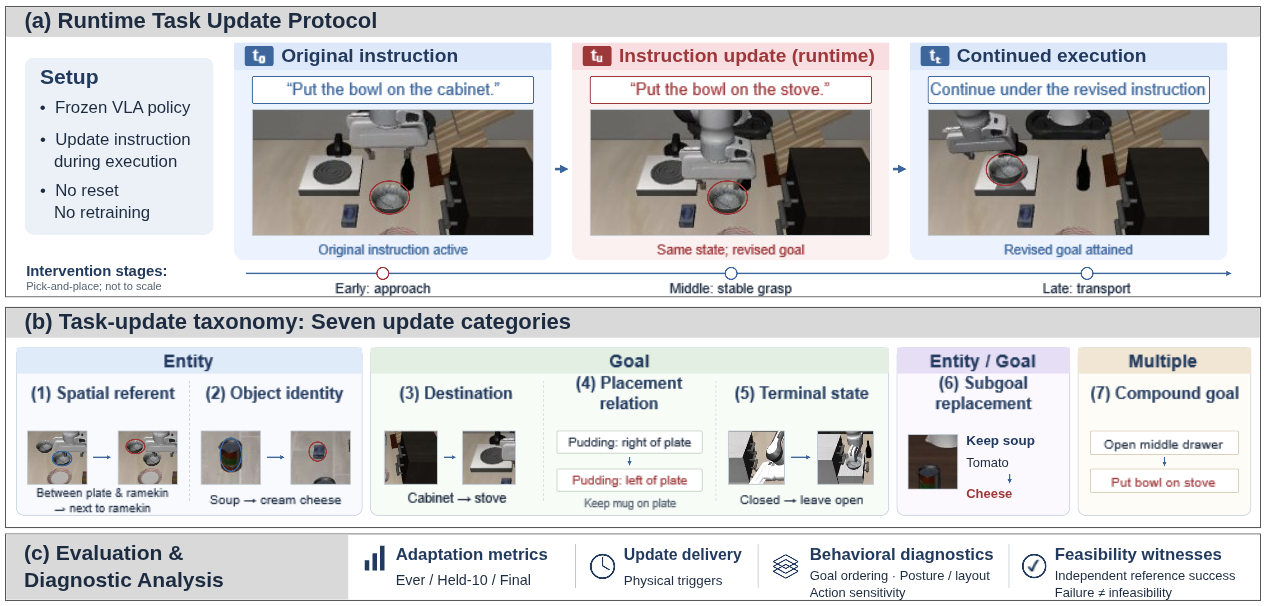}
    \caption{Overview of DTU-Bench.}
    \label{fig:overview}
\end{figure*}

\section{Introduction}
\label{sec:intro}

Robots operating alongside people may need respond to changes in instructions \cite{thierauf2024instead,sharma2022correcting}. A user may change where an object should be placed, request a different object, or withdraw part of an instruction after the robot has already begun acting. Unlike starting a new task, such updates arrive while the robot is executing and may have already made some progress on the initial task. When an instruction change arrives, some actions associated with the initial task may still be useful, while others may need to be abandoned or undone \cite{li2023interactive}. 

Recent advances in robot learning have greatly improved language-conditioned manipulation. Generalist vision-language-action (VLA) policies such as OpenVLA~\cite{kim2024openvla} and $\pi_0$~\cite{black2024pi_0} have demonstrated increasingly capable manipulation across tasks and environments. Benchmarks such as CALVIN~\cite{mees2022calvin} and LIBERO~\cite{libero} have supported evaluation of VLA policies on long-horizon task execution, generalization, and knowledge transfer. However, success on tasks specified before execution does not necessarily mean a policy can adapt when the task changes during execution. A policy may successfully execute the original or the updated tasks independently, yet fail to transition between them once execution has begun.


Recent studies have addressed different challenges in dynamic robot execution. DynamicVLA~\cite{xie2026dynamicvla} focuses on manipulating moving objects, while TIC-VLA~\cite{huang2026tic} addresses reactive control. SwitchVLA~\cite{li2025switchvla} and ReSteer~\cite{chen2026resteer} develop training-based approaches to task switching. However, how existing VLA policies respond when different parts of a task are changed during execution remains insufficiently understood. Moreover, reaching the updated goal does not necessarily indicate that the robot has responded to the new instruction. This motivates a systematic evaluation of runtime task adaptation and post-update behavior.


We introduce Dynamic Task Update Benchmark (\textbf{DTU-Bench}) to systematically evaluate how robot policies respond to task instructions that change during execution. As illustrated in Fig.~\ref{fig:overview}, DTU-Bench introduces new instructions while the robot is acting, without resetting its state. Built on LIBERO, the benchmark covers 40 task revisions across seven types of instruction changes and three execution stages. We evaluate each policy under three conditions: executing the original task, performing the revised task from the initial state of the original task, and switching to the revised task during execution. These comparisons allow us to assess whether policies can complete a revised task after making progress towards a different objective. Beyond task success, we examine whether robots maintain the requested outcome and how their behavior changes following an instruction update.

Our main contributions are:
\begin{itemize}
    \item We introduce DTU-Bench, a systematic, policy-agnostic framework for evaluating runtime task updates in robotic manipulation across seven task-revision categories and three execution stages. The benchmark and evaluation code will be released.

    \item We evaluate 13 VLA configurations and identify limitations that fixed-instruction evaluation can overlook, including differences across update types, delayed satisfaction of updated goals, and substantial gaps between
    temporary and final goal satisfaction.

    
    \item We conduct controlled experiments to examine how the robot's execution state affects responses to new instructions, including whether simple scripted interventions can help recover from failed task updates.

\end{itemize}

\section{Related Work}
\label{sec:related_work}

\noindent\textbf{Embodied Foundation Models.}
Vision--language--action (VLA) models for robotic manipulation map visual observations
and instructions to actions through autoregressive token
generation~\cite{kim2024openvla,pertsch2025fast,wang2026unified},
continuous regression~\cite{kim2025fine}, or flow
matching~\cite{black2024pi_0,intelligence2025pi_}.
WorldVLA and RynnVLA-002 jointly model actions and future
visual observations~\cite{cen2025worldvla,cen2025rynnvla},
while VLA-JEPA and Light-WAM incorporate predictive
representations or supervision~\cite{sun2026vla,li2026light}.
Hierarchical models such as $\tau_0$-VLA further integrate
world-model-guided planning~\cite{cai2026tau_0}.
Despite this architectural diversity, adaptation to
mid-execution task changes remains insufficiently evaluated.

\noindent\textbf{Dynamic Execution and Task Adaptation.}
DynamicVLA~\cite{xie2026dynamicvla} addresses moving-object
manipulation, while TIC-VLA~\cite{huang2026tic} combines
asynchronous reasoning and control. These methods primarily
respond to changing observations rather than task objectives. More directly, SwitchVLA~\cite{li2025switchvla} trains a contact-aware policy for task switching, while ReSteer~\cite{chen2026resteer} improves policy steerability through generated training data. Rather than developing a new switching policy, DTU-Bench focuses on evaluating different types of task revisions, distinguishing revised-task competence from runtime performance and examining whether updated-goal success reflects a response to the new instruction.

\noindent\textbf{Benchmarking VLA Robustness and Adaptation.}
LIBERO~\cite{libero} provides a benchmark for language-conditioned
manipulation tasks; LIBERO-Plus~\cite{fei2026libero}
evaluates robustness to visual, spatial, and linguistic
perturbations; and LIBERO-Safety~\cite{cui2026libero}
examines physical and semantic safety.
DTU-Bench instead focuses on changes to task requirements
during execution. It investigates original tasks,
updated-from-start controls, and runtime updates, assessing
whether updated goals are reached and maintained, alongside
diagnostics of post-update behavior.

\section{DTU-Bench}
\label{sec:dtu_bench}
DTU-Bench (Fig.~\ref{fig:overview}) evaluates runtime task
adaptation in robotic manipulation through structured
task updates, execution-stage interventions, and measures
of updated-goal attainment, persistence, and post-update behavior.
\subsection{Task Update Formulation}
\label{sec:formulation}

In conventional language-conditioned manipulation, a policy executes a fixed instruction $I_0$ toward a goal $G_0$. In DTU-Bench, execution begins under $I_0$, and a revised instruction $I_1$, specifying goal $G_1$, is delivered at time $t=\tau$ while the system is in an intermediate state $s_\tau$. The instruction conditioning the policy changes at the intervention time:
\begin{equation}
 a_t \sim
 \begin{cases}
 \pi(\cdot\mid o_t,I_0), & t<\tau,\\
 \pi(\cdot\mid o_t,I_1), & t\geq\tau,
 \end{cases}
 \label{eq:instruction_switch}
\end{equation}
where $o_t$ is the current observation. Execution of $I_1$ starts from the physical state $s_\tau$ reached under $I_0$ rather than restarting the episode. This distinguishes runtime adaptation from executing the revised task from the initial state: earlier actions may still be useful, may need to be redirected, or may have created conflicts with the new goal.

\subsection{What Changes: Task Update Taxonomy}
\label{sec:taxonomy}

We ground our taxonomy in the task representation of classical planning. Given a fixed planning domain, a planning problem can be described by a set of objects $\mathcal{O}$, an initial state $s_0$, and a goal specification $G$~\cite{fox2003pddl2}:
\begin{equation}
    \mathcal{P} = (\mathcal{O}, s_0, G).
    \label{eq:planning_problem}
\end{equation}
Here, $G$ specifies the conditions required to complete a task. In DTU-Bench, instruction updates change the target object, the required task outcome, or both, while preserving the objects and scene at the time of the update.

We include seven types of task updates, grouped into four broader categories (Table~\ref{tab:taxonomy}):
\emph{Entity (2 types)} changes which object the robot must manipulate (Object identity) or how the target object is identified (Spatial referent);
\emph{Goal (3 types)} changes where an object should be placed (Destination), its required relation to another object (Placement relation), or the required final state of the environment (Terminal state);
\emph{Entity/Goal (1 type)} replaces one required subgoal with another while leaving the remaining subgoals unchanged (Subgoal replacement); and
\emph{Multiple (1 type)} changes two or more independent task requirements within the same instruction (Compound goal).

Each task-update scenario is assigned to one of the seven types, although some changes may involve more than one aspect of the task. Additional experiments on task cancellation and unrelated instructions are conducted on a subset of policies, with results reported in the supplementary material. 


\begin{table}[t]
\centering
\caption{Task-update taxonomy in DTU-Bench.
Categories use source tasks from the indicated LIBERO
suites, with 40 task updates in total.}
\label{tab:taxonomy}

\begingroup
\footnotesize
\setlength{\tabcolsep}{2pt}
\renewcommand{\arraystretch}{1.15}

\begin{tabularx}{\columnwidth}{
  @{}
  >{\raggedright\arraybackslash}p{0.17\columnwidth}
  >{\raggedright\arraybackslash}p{0.38\columnwidth}
  >{\raggedright\arraybackslash}X
  @{}
}
\toprule
\textbf{Dimension} &
\textbf{Update category (LIBERO suite)} &
\textbf{What changes?} \\
\midrule
\multirow[t]{2}{*}{Entity}
 & Spatial referent (Spatial)
 & Relational target description \\
 & Object identity (Object)
 & Manipulated object \\
\midrule
\multirow[t]{3}{*}{Goal}
 & Destination (GOAL)
 & Target location \\
 & Placement relation (Long)
 & Object--destination relation \\
 & Terminal state (Long)
 & Required final condition \\
\midrule
Entity / Goal
 & Subgoal replacement (Long)
 & Required subgoal object \\
\midrule
Multiple
 & Compound goal (GOAL)
 & Multiple task requirements \\
\bottomrule
\end{tabularx}

\endgroup
\end{table}



\subsection{Constructing Task-Update Scenarios from LIBERO}
\label{sec:scenario_construction}

We construct 40 task-update scenarios using tasks from LIBERO-Spatial, Object, GOAL, and Long. Each scenario specifies an original instruction--goal pair $(I_0,G_0)$ and an updated pair $(I_1,G_1)$, both using objects from the same LIBERO scene. The updated instruction changes what the robot is asked to do without modifying the environment. Each scenario is evaluated using the 50 official LIBERO initial states. 



\subsection{When It Changes: Execution-Stage Interventions}
\label{sec:intervention}

We introduce instruction changes at three stages of execution: \emph{early}, \emph{middle}, and \emph{late}. These stages are defined by task-progress events rather than fixed action counts. For example, in pick-and-place tasks, they correspond to approaching an object, grasping or lifting it, and moving it towards its destination. If the robot does not reach the required stage, the update is recorded as not delivered.

When the instruction changes, we discard any remaining actions predicted under the original instruction and generate new actions using the revised instruction and current observations.

The robot is then given 300 action steps to complete the revised task, regardless of how many steps it has already executed. We continue execution even after the revised task is completed to check whether the required outcome is maintained. We additionally evaluate selected continuations over 600 actions to examine how a longer execution horizon affects task completion and maintenance.

\subsection{What Remains: Progress Compatibility and Execution State}
\label{sec:progress}

The effect of a task update depends not only on what changes in the goal, but also on the progress made and the physical state reached during execution prior to the update. We consider two complementary aspects of compatibility with the updated goal $G_1$.

\emph{Task-level progress compatibility} concerns which parts of the original task remain useful, which are no longer needed, and what must be done to complete the revised task. In multi-stage tasks, completed subgoals may remain useful even when other requirements change. For example, if both tasks require opening a drawer, the robot does not need to open it again.


\emph{Execution-state compatibility} concerns how the robot's physical configuration affects subsequent behavior. At the intervention state $s_\tau$, robot posture, existing grasps, and object configurations may facilitate or constrain execution of the updated goal. For example, changing the destination of a grasped object preserves useful progress, whereas changing the target object may require releasing the grasped object and selecting another. 

These two aspects motivate our analysis of post-update behavior (Sec.~\ref{sec:diagnostic_setup}).

\subsection{Evaluation and Diagnostic Analysis}
\label{sec:evaluation}

DTU-Bench evaluates runtime task adaptation through complementary measures of task competence, update delivery, goal attainment, and post-update behavior. Table~\ref{tab:evaluation} summarizes the evaluation components and the evidence each provides. The benchmark protocol applies to all evaluated policies, while controlled diagnostics are performed on selected tasks and configurations.

\textbf{Task competence and update delivery.}
For each task and initial state, we evaluate three conditions:
(1) an \emph{original control} executing $I_0$ from the initial state $s_0$;
(2) an \emph{updated-from-start control} receives $I_1$ before any task actions are executed and starts from the initial state $s_0$; and
(3) \emph{runtime updates} that begin with $I_0$ and switch to $I_1$ at the intervention state $s_\tau$. The first two conditions evaluate performance on each task separately, while the third evaluates performance when the instruction changes during execution. 


\textbf{Task Success and Goal Maintenance.}
We evaluate the updated-goal predicate $G_1$ after each executed action using three metrics: \emph{Ever}, whether the goal is satisfied at least once; \emph{Held-10}, whether it remains satisfied for ten consecutive observations; and \emph{Final}, whether it is satisfied at the end of the 300-step post-update execution. Held-10 measures sustained goal attainment over a short interval, while Final measures whether the goal remains satisfied at the end of execution. Their difference helps identify cases where the updated goal is temporarily attained but not maintained until the end of execution.

\textbf{Matched-state behavioral diagnostics.}
Goal attainment alone does not establish whether a policy changes its behavior in response to an updated instruction. For selected intervention states, we restore the same recorded state and compare two continuations: one retaining the original instruction $I_0$ and the other receiving the updated instruction $I_1$. Both branches use the same updated-goal predicate $G_1$ and an equal action budget. We compare their goal-attainment outcomes, object interactions, and the temporal order of original- and updated-goal satisfaction. This design tests whether attaining $G_1$ depends on receiving the updated instruction under matched execution conditions.



\textbf{Execution-state and attainability diagnostics.}
We examine how robot posture, object layout, and visual observations affect post-update behavior. A separate geometry-based reference controller uses simulator object positions and predefined waypoints to independently attempt the updated task (details in the supplementary material). Its success demonstrates attainability, while failure does not establish infeasibility.

\begin{table}[t]
\centering
\caption{DTU-Bench evaluation components.
Standard benchmark measures quantify task competence
and goal satisfaction; controlled diagnostics examine
post-update behavior and attainability.}
\label{tab:evaluation}

\begingroup
\footnotesize
\setlength{\tabcolsep}{4pt}
\renewcommand{\arraystretch}{1.12}

\begin{tabularx}{\columnwidth}{
  @{}
  >{\raggedright\arraybackslash}p{0.35\columnwidth}
  >{\raggedright\arraybackslash}X
  @{}
}
\toprule
\textbf{Component} &
\textbf{What it establishes} \\
\midrule
\multicolumn{2}{@{}l}{\textit{Standard benchmark evaluation}} \\
Original control &
Original-task competence \\
Updated-from-start &
Updated-task competence \\
Runtime update &
Updated-goal attainment from $s_\tau$ \\
Update delivery &
Whether the intervention occurs \\
Held-10 &
Ten consecutive observations satisfying $G_1$ \\
Final &
Satisfaction of $G_1$ at execution end \\
\midrule
\multicolumn{2}{@{}l}{\textit{Controlled diagnostic analysis}} \\
Matched continuations &
Instruction-dependent outcomes from the same state \\
Execution-state intervention &
Sensitivity to posture, layout, and observations \\
Reference controller &
Attainability evidence for selected states \\
\bottomrule
\end{tabularx}
\endgroup
\end{table}

\section{Experiments}
\label{sec:experiments}

\subsection{Experimental Setup}
\label{sec:exp_setup}

\textbf{Benchmark and evaluation protocol.}
We evaluate DTU-Bench using all four LIBERO suites: Spatial, Object, GOAL, and Long. The full evaluation design includes 40 source tasks, 50 official initial states per task, and five execution conditions: original-from-start, updated-from-start, and runtime updates at early, middle, and late stages. The full evaluation design comprises 10,000 scheduled trials per policy configuration, with the two from-start controls shared across the three intervention stages. Each delivered update is evaluated for 300 executed actions.


We report the evaluation coverage, valid trials, and exclusions for each policy in the supplementary material. Trials with simulator errors are excluded from the success-rate calculations.

\textbf{VLA baselines.}
We evaluate a series of publicly released VLA configurations covering different action-generation mechanisms (autoregressive action tokens, direct continuous regression, and flow-based action prediction). The models also span conventional vision-language-action architectures, action-expert policies, and approaches incorporating world-model or predictive video supervision:
(1) OpenVLA~\cite{kim2024openvla},
(2) OpenVLA-OFT~\cite{kim2025fine},
(3) $\pi_0$~\cite{black2024pi_0},
(4) $\pi_{0.5}$~\cite{intelligence2025pi_},
(5) DM0.5 \cite{dm05},
(6) WorldVLA~\cite{cen2025worldvla},
(7) RynnVLA-002~\cite{cen2025rynnvla},
(8) UniVLA~\cite{wang2026unified},
(9) VLA-JEPA~\cite{sun2026vla},
(10) InstructVLA~\cite{yang2026vision},
(11) $\tau_0$-VLA~\cite{cai2026tau_0},
(12) MolmoAct2~\cite{fang2026molmoact2}, and
(13) FlashVLA-$\pi_{0.5}$~\cite{li2026flashvla}.

We use released LIBERO checkpoints without additional training. Each configuration retains its native observation inputs, action representation, and inference interface while following the same task-update protocol and goal evaluation. For models with additional planning or world-model components, we evaluate only the released policy configuration supported by the tested interface, rather than assuming that the full planning system is active. Checkpoint details, execution settings, and evaluation coverage are provided in supplementary material.

\textbf{Statistical analysis.}
We estimate pointwise 95\% confidence intervals by resampling initial-state
blocks within each source task, keeping all conditions and intervention
stages paired. Task-cluster bootstrap and leave-one-task-out analyses assess
sensitivity to task composition. Full results are reported
in the supplementary material.

\subsection{Controlled Diagnostic Experiments}
\label{sec:diagnostic_setup}



We conduct additional diagnostics on selected LIBERO tasks using recorded intervention states and fixed policy checkpoints. These experiments examine specific benchmark behaviors rather than estimate performance across all policies or task categories.

\textbf{Destination versus object updates.}
Using $\pi_0$ and $\pi_{0.5}$, we compare destination ($A{\rightarrow}P$ to $A{\rightarrow}Q$) and object ($A{\rightarrow}P$ to $B{\rightarrow}P$) updates in the same LIBERO-GOAL scene. We test five initial states at approach and stable-grasp stages with 300-action continuations. 

\textbf{Instruction and execution configuration.}
For $\pi_{0.5}$ and DM0.5, we vary instruction (original/updated), posture (saved/neutral), and object layout (original/swapped) across two LIBERO-Object tasks and five initial states, yielding 160 continuations. Neutral-posture interventions preserve object positions within each layout. We measure first-object contact and
updated-goal success over 600 actions.


\textbf{Visual-input sensitivity.}
We examine how external- and wrist-camera observations affect action predictions under the original and updated instructions. The experimental setup and full results are provided in the supplementary material.



\textbf{Trajectory analysis.} We automatically extract first finger--object contact and the times at which the original and updated task conditions are satisfied from simulator logs. Shared and obsolete task requirements are predefined for each scenario and evaluated automatically after each executed action. These records allow us to identify whether the original task is completed before the updated task, or whether the updated instructions are satisfied only temporarily. For selected 600-action trajectories, we also evaluate outcomes at 300 actions to compare different execution horizons using the same trajectory.

\textbf{Matched instruction continuations.} We compare original- and updated-instruction continuations from identical restored intervention states, evaluating both against the same updated-goal predicate $G_1$. The 300-action diagnostic contains 94 matched pairs (46 for $\pi_{0.5}$ and 48 for DM0.5) across selected LIBERO-Long, Object, and Spatial tasks. Both branches use equal continuation budgets, with pending actions discarded before fresh inference. Additional suffix comparisons reuse original branches and are not treated as independent matched pairs.

\textbf{Inference-time assistance.}
We investigate whether simple training-free helper steps can recover failed updates. For $\pi_{0.5}$ and DM0.5, we compare three methods from the same recorded intervention states:
\emph{Direct}, which immediately executes the updated instruction;
\emph{Retreat}, which first moves the arm to a predefined joint configuration (robot default initial pose); and
\emph{Release+Retreat}, which opens the gripper before retreating when it is closed.
The helpers use only robot joint and gripper information without accessing object positions, contact labels, or task-success conditions to control their actions. We evaluate all ten LIBERO-Object tasks and five LIBERO-GOAL destination-update tasks, using five initial states and two intervention stages (approach and stable grasp). All methods share the same updated instruction and 300-action budget, including assistance. 

\begin{table*}[t]
\centering
\caption{\textbf{Original-task and runtime updated-goal success.}
Each cell reports Original Final $\rightarrow$ Runtime Final (\%).
Darker green indicates higher runtime success on a common 0--100 scale;
bold marks the highest reported runtime score in each column.}
\label{tab:dtu_main_final}

\begingroup
\footnotesize
\setlength{\tabcolsep}{2pt}
\renewcommand{\arraystretch}{1.15}

\begin{tabular*}{\textwidth}{
  @{\extracolsep{\fill}}l*{7}{c}@{}
}
\toprule
Policy
& \shortstack{Spatial\\referent}
& \shortstack{Object\\identity}
& Destination
& \shortstack{Placement\\relation}
& \shortstack{Terminal\\state}
& \shortstack{Subgoal\\replacement}
& \shortstack{Compound\\goal} \\
\midrule

OpenVLA
& \DTUcell{0}{81.2}{0.0}
& \DTUcell{0}{85.4}{0.0}
& \DTUcell{36.4}{76.8}{36.4}
& \DTUcell{1.8}{24.2}{\mathbf{1.8}}
& \DTUcell{5.6}{34.0}{\mathbf{5.6}}
& \DTUcell{0}{63.5}{0.0}
& \DTUcell{6.8}{67.2}{6.8} \\

OpenVLA-OFT
& \DTUcell{0}{96.4}{0.0}
& \DTUcell{0}{98.0}{0.0}
& \DTUcell{45.3}{92.8}{45.3}
& \DTUcell{0}{55.3}{0.0}
& \DTUcell{3.1}{85.3}{3.1}
& \DTUcell{0}{90.5}{0.0}
& \DTUcell{0.9}{90.0}{0.9} \\

$\pi_{0.5}$
& \DTUcell{0}{95.8}{0.0}
& \DTUcell{0.7}{98.0}{0.7}
& \DTUcell{68.8}{94.4}{\mathbf{68.8}}
& \DTUcell{0}{50.3}{0.0}
& \DTUcell{0.7}{74.7}{0.7}
& \DTUcell{0.2}{96.0}{0.2}
& \DTUcell{5.3}{90.8}{5.3} \\

DM0.5
& \DTUcell{0.1}{98.4}{0.1}
& \DTUcell{13.5}{99.8}{\mathbf{13.5}}
& \DTUcell{57.2}{95.6}{57.2}
& \DTUcell{1.1}{39.6}{1.1}
& \DTUcell{0.2}{90.0}{0.2}
& \DTUcell{0.2}{98.0}{0.2}
& \DTUcell{17.5}{92.0}{17.5} \\

$\pi_0$
& \DTUcell{0.5}{94.4}{0.5}
& \DTUcell{0}{97.6}{0.0}
& \DTUcell{57.1}{84.8}{57.1}
& \DTUcell{0}{59.7}{0.0}
& \DTUcell{1.3}{62.0}{1.3}
& \DTUcell{0}{70.5}{0.0}
& \DTUcell{18.9}{89.6}{18.9} \\

RynnVLA-002 cont.
& \DTUcell{0}{90.2}{0.0}
& \DTUcell{0.1}{94.8}{0.1}
& \DTUcell{42.8}{91.6}{42.8}
& \DTUcell{0}{61.3}{0.0}
& \DTUcell{0.7}{76.0}{0.7}
& \DTUcell{0}{94.5}{0.0}
& \DTUcell{2.3}{78.0}{2.3} \\

VLA-JEPA
& \DTUcell{0}{96.4}{0.0}
& \DTUcell{0}{99.8}{0.0}
& \DTUcell{57.3}{98.0}{57.3}
& \DTUcell{0}{89.9}{0.0}
& \DTUcell{0.4}{82.7}{0.4}
& \DTUcell{0}{97.5}{0.0}
& \DTUcell{15.6}{92.0}{15.6} \\

InstructVLA stage-1
& \DTUcell{0}{94.6}{0.0}
& \DTUcell{0.1}{99.2}{0.1}
& \DTUcell{53.5}{97.6}{53.5}
& \DTUcell{0}{70.5}{0.0}
& \DTUcell{1.6}{80.0}{1.6}
& \DTUcell{0}{88.5}{0.0}
& \DTUcell{2.7}{91.2}{2.7} \\

$\tau_0$-VLA low-level
& \DTUcell{0.1}{97.0}{0.1}
& \DTUcell{0.1}{99.2}{0.1}
& \DTUcell{54.0}{95.6}{54.0}
& \DTUcell{0}{43.0}{0.0}
& \DTUcell{0}{90.7}{0.0}
& \DTUcell{0}{89.5}{0.0}
& \DTUcell{7.9}{87.6}{7.9} \\

MolmoAct2 cont.
& \DTUcell{0.1}{97.0}{0.1}
& \DTUcell{4.9}{99.0}{4.9}
& \DTUcell{58.1}{97.2}{58.1}
& \DTUcell{1.6}{47.7}{1.6}
& \DTUcell{0.9}{74.7}{0.9}
& \DTUcell{0}{95.0}{0.0}
& \DTUcell{19.5}{92.8}{\mathbf{19.5}} \\

FlashVLA-$\pi_{0.5}$ sync.
& \DTUcell{0.3}{97.2}{0.3}
& \DTUcell{1.3}{99.0}{1.3}
& \DTUcell{57.6}{94.0}{57.6}
& \DTUcell{0}{59.7}{0.0}
& \DTUcell{0}{83.3}{0.0}
& \DTUcell{1.7}{98.0}{\mathbf{1.7}}
& \DTUcell{18.9}{91.6}{18.9} \\

WorldVLA
& \DTUcell{1.2}{69.6}{\mathbf{1.2}}
& \DTUcell{0.6}{79.0}{0.6}
& \DTUcell{38.6}{78.4}{38.6}
& \DTUcell{0}{28.7}{0.0}
& \DTUcell{1.1}{30.7}{1.1}
& \DTUcell{0}{44.5}{0.0}
& \DTUcell{3.9}{67.6}{3.9} \\

UniVLA
& \DTUcell{0}{98.0}{0.0}
& \DTUcell{0}{96.0}{0.0}
& \DTUcell{46.7}{60.0}{46.7}
& \DTUcell{0}{66.7}{0.0}
& \DTUcell{2.2}{73.3}{2.2}
& \DTUcell{0}{85.0}{0.0}
& \DTUcell{5.3}{84.0}{5.3} \\

\bottomrule
\end{tabular*}
\endgroup
\end{table*}

\begin{figure}
    \centering
    \includegraphics[width=\linewidth]{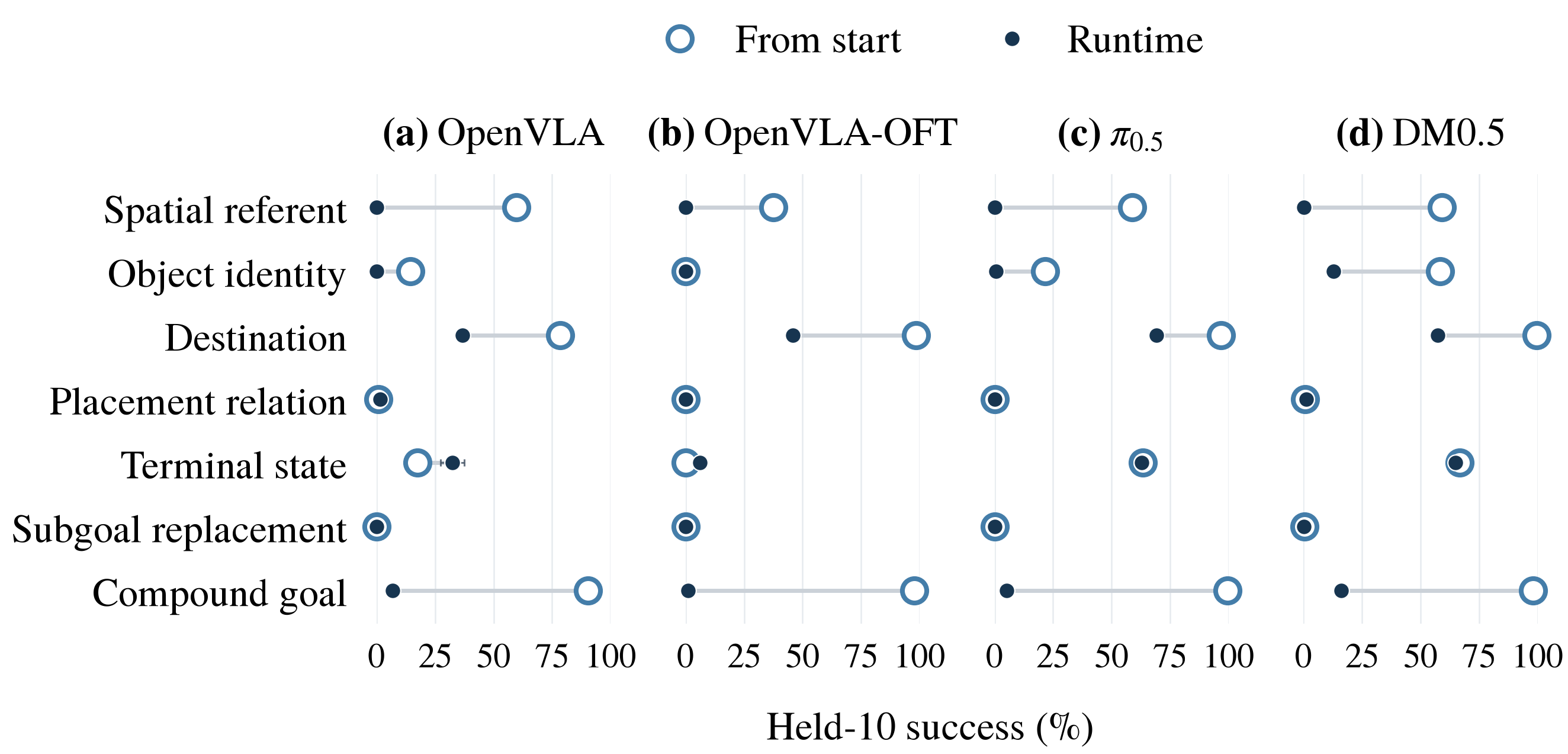}
    \caption{Updated-from-start vs. runtime Held-10 success for four representative VLA policies. Open circles indicate updated-from-start success, filled circles indicate runtime success, and connecting lines show the difference. Results for all 13 policies are provided in the supplementary material.}
    \label{fig:competence_runtime}

\end{figure}

\section{Results}
\label{sec:results}

We first compare performance across task updates and intervention stages. We then examine what robots do after an update, using matched continuations, object-selection tests, and reference trajectories.

\subsection{Responses Differ by Task Update Types}
\label{sec:results_overview}

Table~\ref{tab:dtu_main_final} compares the final task success of original-task and runtime updated-goal across seven update categories. Because these scores measure different task requirements, their difference cannot be attributed solely to the instruction change. Across the reported configurations, destination updates resulted in 36.4--68.8\% runtime Final success, compared with at most 13.5\% yielded by object-identity updates and 1.2\% for spatial-referent updates. Destination changes often preserve manipulation of the same object, whereas object updates require selecting another target. However, scene layout and manipulation task difficulty also vary across categories, preventing attribution of the observed performance difference solely to the type of task update. We confirm this gap in a same-scene comparison (Sec.~\ref{sec:results_behavior}).

\noindent\textbf{Finding 1:}
The VLAs achieved substantially higher task success rates when encountering destination updates compared to object-identity and spatial-referent updates. However, this difference may be caused by factors in addition to the update type.

\begin{figure}
    \centering
    \includegraphics[width=\linewidth]{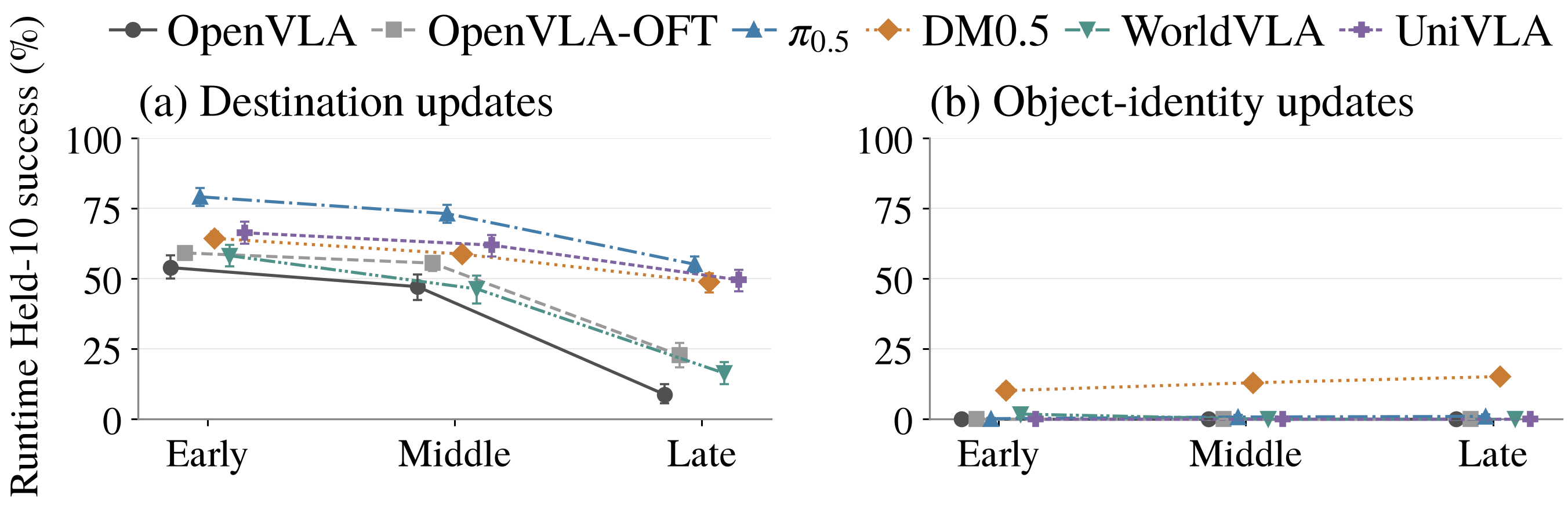}
    \caption{Effect of intervention stage on runtime success. Held-10 success for two update categories and three intervention stages for six VLA policies. Rates are computed over valid scheduled trials, including updates that were not delivered. Error bars show 95\% CIs. Full 13-policy results are in the supplementary material.}
    \label{fig:intervention_stage}
\end{figure}

\subsection{Task Competence $\ne$ Runtime Success}
\label{sec:results_competence}

A failed runtime update does not necessarily mean the system failed to respond to the new instruction. It may be due to the difficulty of the updated task itself. 
We compare Held-10 success across 13 VLA configurations when the updated instruction is given before execution begins versus during execution, after the robot has started the original task. Figure~\ref{fig:competence_runtime} highlights four representative policies: OpenVLA, OpenVLA-OFT, $\pi_{0.5}$, and DM0.5. Across all 13 configurations, runtime Held-10 success is often substantially lower than updated-from-start success, although the gaps vary across update categories. For example, OpenVLA-OFT shows a 96.7 pp gap on compound-goal updates (95\% CI: 94.5--98.4 pp). 
This reveals a gap between performing a task in isolation and achieving the same goal after partial execution under a different instruction.


\noindent\textbf{Finding 2:} Success on an updated task from the start does not guarantee success when the same instruction is introduced during execution.


\subsection{Intervention Stage Affects Runtime Success}
\label{sec:results_timing}

We examine how intervention stage affects runtime Held-10 success across all 13 evaluated policy configurations. Figure~\ref{fig:intervention_stage} shows six representative policies, illustrating different patterns across update types. 
For destination updates, success generally decreases from early to late intervention. For example, DM0.5 drops from 64.4\% to 58.8\% and 48.8\%. In contrast, most policies achieve near-zero success on object-identity updates across all stages, except DM0.5, which shows a slight increase.
These results suggest that the effect of intervention stage depends on the task requirements being changed, rather than following a uniform early-to-late decline.

\noindent\textbf{Finding 3:}
Later interventions are not consistently harder; their effect on runtime success varies across task-update types.

\subsection{Reaching the New Goal $\ne$ Adaptation Success}
\label{sec:results_persistence}

Runtime success measures whether the updated goal is achieved, but does not capture how or when it is reached. 

\textbf{Instruction-dependent goal attainment.} We compare 94 matched pairs of continuations from identical intervention states, with one branch following the original instruction and the other receiving the update. Both branches are evaluated against the updated goal. Under Held-10, 37 pairs succeed with both instructions, three succeed only with the update, and 54 fail under both. 
All 37 both-success cases involve terminal-state updates, indicating that the updated goal can be temporarily satisfied even when following the original instruction. However, at the end of execution (Final), only five updated-instruction continuations still satisfy the updated goal, while none of the original-instruction continuations do.

\begin{figure}
    \centering
    \includegraphics[width=1\linewidth]{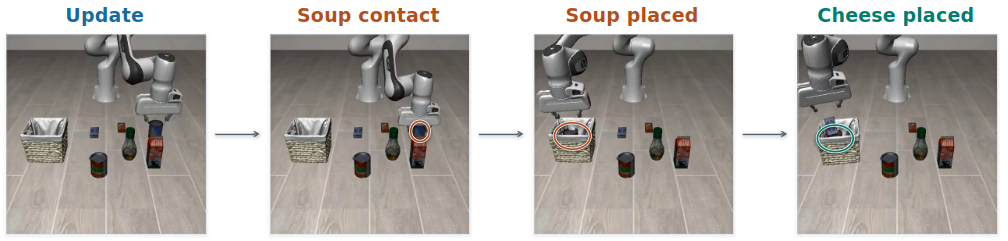}
    \caption{Example DM0.5 trajectory in LIBERO-Object, where the instruction changes from alphabet soup to cream cheese. The frames show the robot contacting and placing the original object before completing the revised goal.} 
    \label{fig:delayed_completion}
\end{figure}

\textbf{Delayed goal completion.}
In a LIBERO-Object task where the instruction changes from placing alphabet soup to placing cream cheese, DM0.5's Held-10 success increases from 4/15 at 300 actions to 15/15 at 600 actions. All 15 trajectories complete the original goal before the updated goal (Fig.~\ref{fig:delayed_completion}). 
This behavior extends across multiple VLAs: among successful Object updates, the original goal is completed first in 95.8\% of DM0.5 cases (184/192), 90.8\% for MolmoAct2 (59/65), 94.7\% for FlashVLA (18/19), and 90.0\% for $\pi_{0.5}$ (9/10). These results show that achieving the updated goal does not necessarily indicate a prompt response to the instruction change.

\textbf{Temporary goal satisfaction.}
Terminal-state updates provide another example of why goal attainment alone can be misleading. For terminal-state updates, DM0.5 achieves 64.9\% Held-10 success but only 0.22\% Final success, while $\pi_{0.5}$ achieves 62.9\% and 0.67\%, respectively. For example, a robot may temporarily leave a drawer open before continuing to close it. Such outcomes distinguish temporary goal attainment from sustained satisfaction.

\textbf{Preserving useful progress.}
Runtime updates do not always require abandoning previous work. 
In a LIBERO-Long example, the instruction changes from placing cream cheese and butter in a basket to placing cream cheese and soup in the basket. DM0.5 preserves the already completed cheese placement and adds the newly requested soup without placing the obsolete butter.

\noindent\textbf{Finding 4:}
Reaching the updated goal does not necessarily mean the robot changed its behavior promptly or maintained the goal afterward.

\begin{figure}
    \centering
    \includegraphics[width=0.8\linewidth]{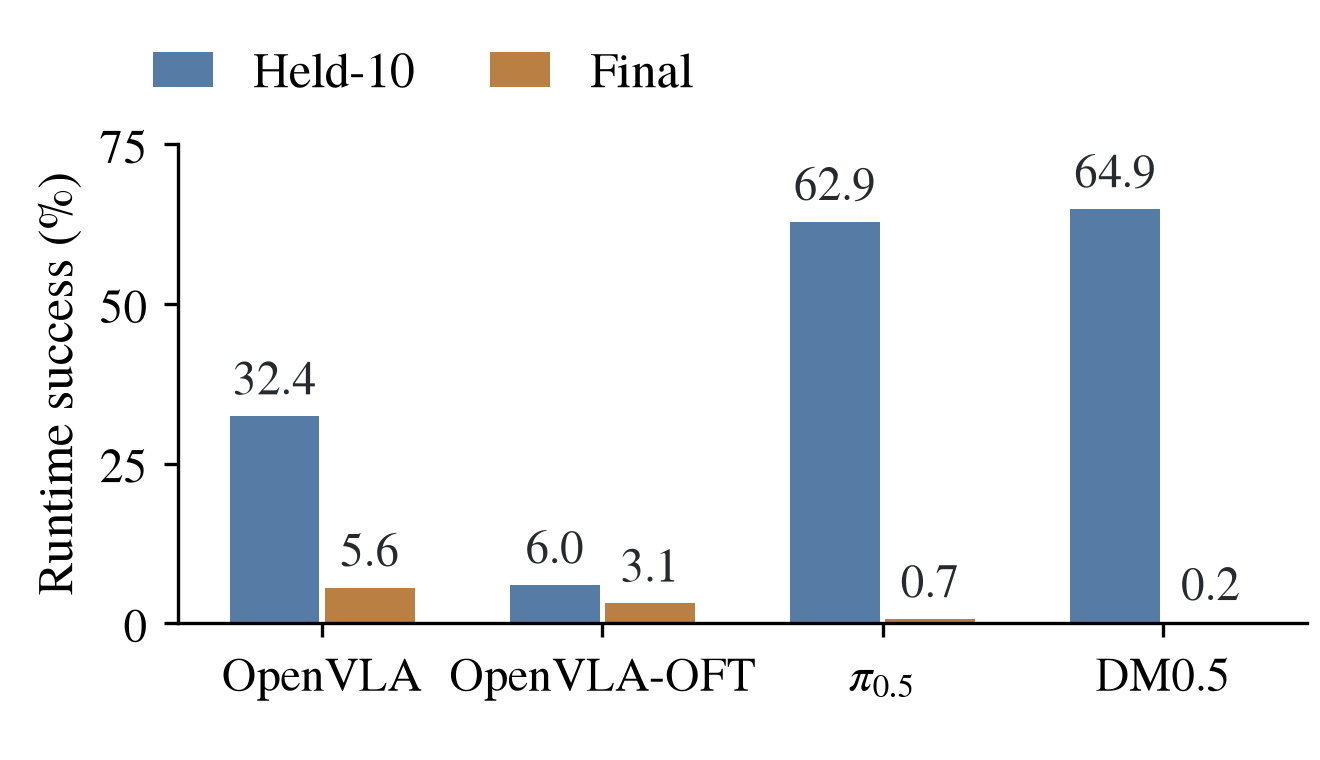}
    \caption{Held-10 and Final success for LIBERO-Long terminal-state updates. Each policy has 450 scheduled trials (3 tasks $\times$ 50 states $\times$ 3 intervention stages). Held-10 requires the revised goal to hold for 10 consecutive executed-action observations; Final requires it to hold at the last observation. Full 13-policy results are in the supplementary material.}

    \label{fig:terminal_persistence}
\end{figure}

\subsection{Task Context Affects Post-Update Behavior}
\label{sec:results_behavior}

Updated-goal success alone does not reveal how policies respond to instruction changes from different execution states. We therefore conduct controlled experiments to examine how prior task progress and robot posture affect subsequent behavior.

\textbf{Destination versus object updates.}
To examine whether partial execution affects different types of updates, we compare destination and object changes from matched intervention stages using $\pi_0$ and $\pi_{0.5}$. Both policies succeed in all 10 destination-update continuations, including after grasping the original object. In contrast, neither succeeds in the five object-update continuations after stable grasp, although all updated-from-start controls succeed. Before grasping, object-update success is 0/5 for $\pi_0$ and 3/5 for $\pi_{0.5}$. 
These results show that the gap between destination and object updates (Sec.~\ref{sec:results_overview}) persists when scene and task difficulty are held fixed: progress made under the original instruction remains useful for destination changes but can hinder updates requiring a different object.

\begin{figure}
    \centering
    \includegraphics[width=1\linewidth]{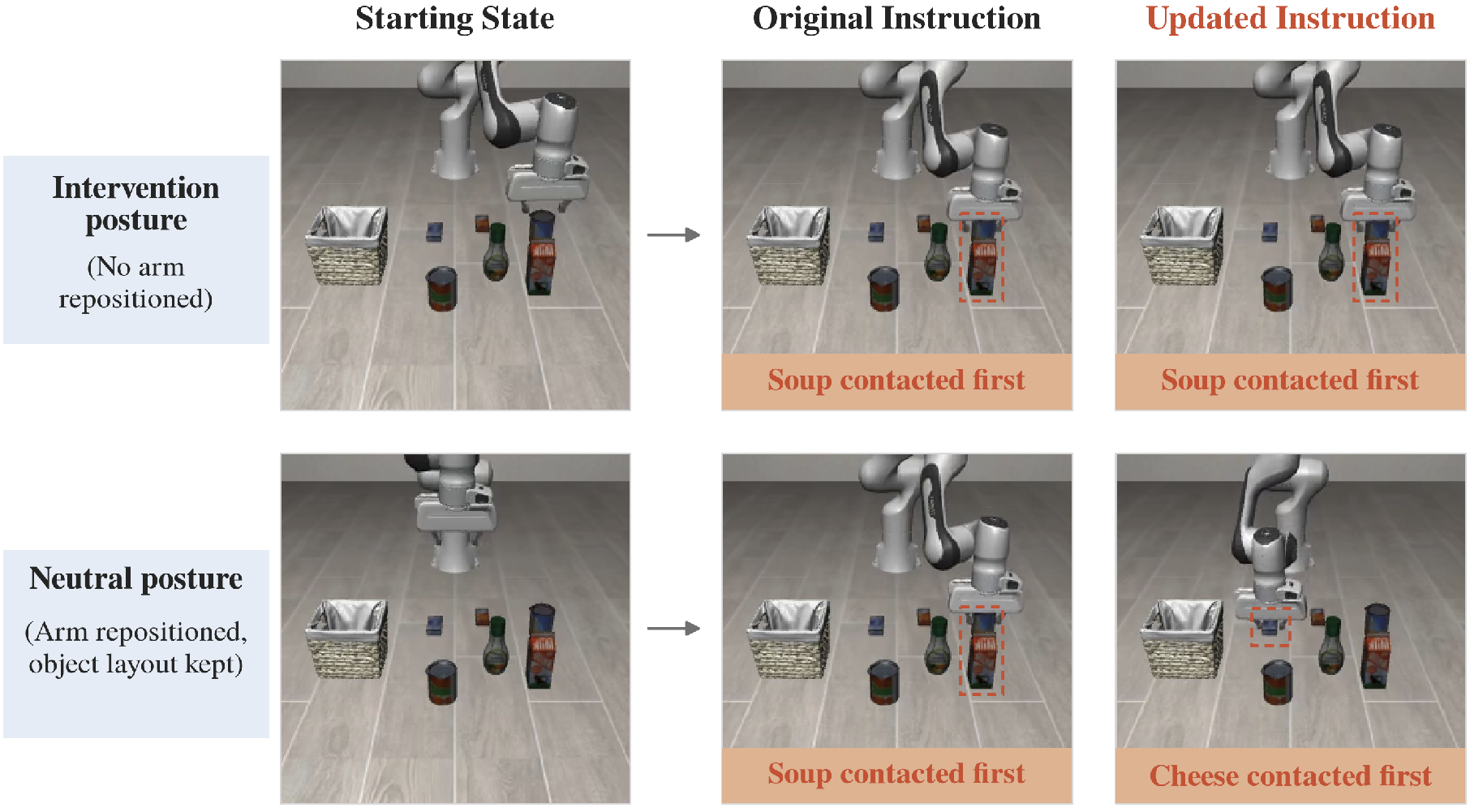}
    \caption{Object selection under different robot postures and instructions. DM0.5 continuations under original (soup) and updated (cheese) instructions at the saved intervention posture (top) and neutral posture (bottom), with object layout unchanged. Red dashed boxes mark the first contacted object. Results are consistent across five tested states per condition.}

    \label{fig:instruction_posture}
\end{figure}

\textbf{Instruction, posture, and object layout.}
We next investigate how the robot's current configuration affects object selection. In 160 matched continuations with $\pi_{0.5}$ and DM0.5, we vary the instruction, robot posture, and object layout across two Object source tasks. Figure~\ref{fig:instruction_posture} illustrates how DM0.5 selects different objects under the same updated instruction when the arm posture changes. At the saved intervention posture, both policies first contact the nearer candidate object in all 40 tested conditions per policy, while only 20/40 contacts match the specified instruction. Swapping the object layout changes which object is contacted first. At the neutral posture, instruction-consistent first contact increases to 40/40 for DM0.5 and 29/40 for $\pi_{0.5}$. The results therefore differ between policies. Moreover, even when DM0.5 contacts the updated target first in all 20 relevant neutral-posture continuations, only 18/20 achieve the updated goal at the final evaluation step. These results show that selecting the requested object and completing the updated task are different outcomes.

\textbf{Attainability of failed updates.} We use a separate geometry-based controller (Sec.~\ref{sec:evaluation}) that uses simulator object positions and predefined waypoints to complete the updated task without a VLA policy. Across 20 Object-task tests, it succeeds in 10 cases, nine of which fail under the evaluated VLAs. Conversely, DM0.5 succeeds in three cases where the reference controller fails. Thus, some VLA failures occur despite achievable updated tasks, while controller failure does not establish infeasibility.

\noindent\textbf{Finding 5: }
Prior progress and robot state can facilitate or hinder responses to task updates.

\subsection{Assistance to Improve Task Update Success}
\label{sec:results_assistance}

\begin{figure}
    \centering
    \includegraphics[width=0.8\linewidth]{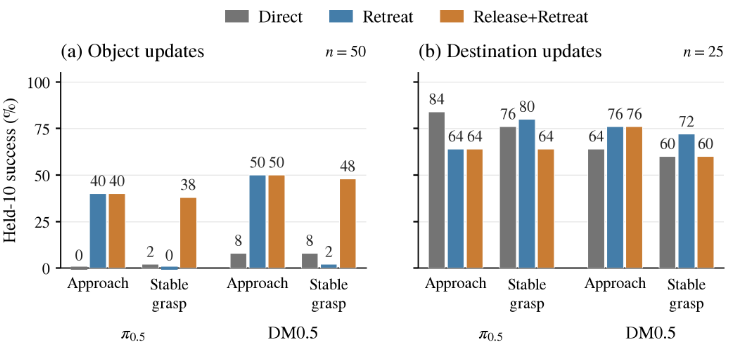}
    \caption{Training-free assistance for task updates.}
    \label{fig:assist}
\end{figure}

The preceding experiments suggest that the robot's physical state affects its response to instruction changes. We therefore test whether simple training-free assistance can improve task completion without retraining the policy (Fig.~\ref{fig:assist}). For each policy and intervention stage, we evaluate 50 Object trials (10 tasks $\times$ 5 states) and 25 Destination trials (5 tasks $\times$ 5 states).

\textbf{Object updates.}
Before grasping, fixed-joint retreat (i.e., return to the robot original pose) increases Held-10 success from 0/50 to 20/50 for $\pi_{0.5}$ and from 4/50 to 25/50 for DM0.5. After stable grasp, however, retreat alone provides little benefit (1/50 to 0/50 and 4/50 to 1/50, respectively). Releasing the object before retreating instead increases stable-grasp success to 19/50 for $\pi_{0.5}$ and 24/50 for DM0.5. Relative to direct updates, the paired improvements are 36 percentage points (95\% task-cluster bootstrap CI: [12, 62]) and 40 percentage points ([18, 64]), respectively. The improvements remain positive when any single Object task is excluded.

\textbf{Destination updates.}
The same assistance does not consistently improve tasks that retain the original target object. For $\pi_{0.5}$, retreat decreases approach-stage success from 21/25 to 16/25, while release followed by retreat decreases stable-grasp success from
19/25 to 16/25. Results for DM0.5 are mixed. These outcomes suggest that unnecessary physical
interventions can disrupt progress that remains useful for the updated task.

\textbf{Physical side effects.}
Although release followed by retreat improves completion of some object updates, unsupported downward object motion occurs in 49/50 stable-grasp trials for each policy. Thus, improved task success does not necessarily imply desirable physical execution. The interventions are performed by external scripted controllers rather than learned recovery behavior of the VLA policies.

\noindent\textbf{Finding 6:}
Simple physical assistance can mitigate some task-update failures without policy retraining, although its benefits and physical consequences depend on the update type and execution stage.


\section{Conclusion}
\label{sec:conclusion}


We introduced DTU-Bench to evaluate how VLA policies respond to instruction changes during execution. Across 13 configurations, we found substantial differences across update types and execution stages, while success on fixed tasks did not guarantee runtime success. Behavioral diagnostics revealed delayed completion and temporary satisfaction of updated tasks. Training-free assistance improved some failures but could also disrupt useful progress and introduce undesirable object motion. These findings highlight the importance of evaluating task success alongside post-update behavior and physical execution.
\section*{LLM Usage Statement}
ChatGPT and OpenAI Codex (OpenAI) were used to assist with manuscript wording, experimental code development and debugging, statistical analysis, and figure formatting and layout adjustments. The authors are responsible for the experimental methodology, results, and scientific conclusions.
{
    \small
    \bibliographystyle{ieeenat_fullname}
    \bibliography{main}
}


\end{document}